\documentclass[conference]{IEEEtran}
\IEEEoverridecommandlockouts
\usepackage{cite}
\usepackage{amsmath,amssymb,amsfonts}
\usepackage{algorithm}
\usepackage{algorithmic}
\usepackage{graphicx}
\usepackage{textcomp}
\usepackage{xcolor}
\usepackage{multicol}
\usepackage{multirow}
\usepackage{subcaption}
\usepackage{mathrsfs}
\usepackage{booktabs}
\usepackage{tikz}
\usepackage{soul}
\usepackage{rotating}
\begin{document}

\title{DyLoc: Dynamic Localization for Massive MIMO Using Predictive Recurrent Neural Networks }

%DyLoc: How Massive MIMO Can Pave The Way For Dynamic Fingerprinting-based Localization Platforms?}
\author{\IEEEauthorblockN{ Farzam Hejazi}
\IEEEauthorblockA{\textit{Department of Electrical}\\ \textit{and Computer Engineering} \\
\textit{University of Central Florida}\\
Orlando, USA \\
farzam.hejazi@ucf.edu}
\and
\IEEEauthorblockN{ Katarina Vuckovic}
\IEEEauthorblockA{\textit{Department of Electrical}\\ \textit{and Computer Engineering}  \\
\textit{University of Central Florida}\\
Orlando, USA \\
kvuckovic@knights.ucf.edu}
\and
\IEEEauthorblockN{ Nazanin Rahnavard}
\IEEEauthorblockA{\textit{Department of Electrical}\\ \textit{and Computer Engineering}  \\
\textit{University of Central Florida}\\
Orlando, USA \\
nazanin.rahnavard@ucf.edu}
\vspace{-20mm}
}

\maketitle
\begin{abstract}
%\NR{first mention what is the goal? then mention the approach.}
This paper presents a data-driven localization framework with high precision in time-varying complex multipath environments, such as dense urban areas and indoors, where GPS and model-based localization techniques come short.
We consider the angle-delay profile (ADP), a linear transformation of channel state information (CSI), in massive MIMO systems and show that ADPs preserve users' motion when stacked temporally. We discuss that given a static environment, future frames of ADP time-series are predictable employing a video frame prediction algorithm. We express that a deep convolutional neural network (DCNN) can be employed to learn the background static scattering environment. To detect foreground changes in the environment, corresponding to path blockage or addition, we introduce an algorithm taking advantage of the trained DCNN. Furthermore, we present \emph{DyLoc}, a data-driven framework to recover distorted ADPs due to foreground changes and to obtain precise location estimations. We evaluate the performance of DyLoc in several dynamic scenarios employing DeepMIMO dataset \cite{alkhateeb2019deepmimo} to generate geo-tagged CSI datasets for indoor and outdoor environments. We show that previous DCNN-based techniques fail to perform with desirable accuracy in dynamic environments, while DyLoc pursues localization precisely. Moreover, simulations show that as the environment gets richer in terms of the number of multipath, DyLoc gets more robust to foreground changes.
\end{abstract}


\section{Introduction}
\label{intro}
With the expansion of location-based services such as peer-to-peer ride sharing, local search-and-discovery mobile apps, navigation services, store locators, autonomous driving, and urban unmanned aerial systems (UAS) traffic management, the demand for highly-accurate positioning technologies is growing \cite{junglas2008location}. When the environment is free from strong multipath (mainly outdoor environments), localization can be considered mostly a solved problem \cite{5208736}. Nonetheless, localization in harsh multipath environments (mainly indoors and dense urban areas) has remained under extensive investigation \cite{8692423}. % Currently, the Global Positioning System (GPS) is the most common positioning system that reaches accuracy levels in the order of several meters in the open air. However, GPS fails to work with admissible precision in indoor and dense urban areas due to the complex multipath environment \cite{GPSacccuracy1} and the need for sub-meter positioning accuracies especially for indoors.
 %The Environmental Protection Agency (EPA) reports that we spend $86.9\%$ of our time in indoor areas such as residences, marketplaces, offices, and restaurants, $5.5\%$ in vehicles and $7.6\%$ outdoors~\cite{klepeis2001national}. Accordingly, most of the time individuals do not have access to a reliable and precise localization system and this becomes more noticeable at airports, large shopping malls, big and multi-level parking garages, downtown in big cities, etc. %, where sub-meter positioning accuracy is required. %Moreover, current autonomous vehicular technology requires sub-meter positioning accuracy for automated overtake, cooperative collision avoidance, and high density platooning which are not on hand using GPS \cite{wymeersch20175g}. 
%Poor performance in \emph{elevation} estimation is another deficiency of GPS. The vertical error of GPS is $3$ fold of its horizontal error \cite{GPSacccuracy}, making it unserviceable for applications that need 3D positioning information especially in urban areas, such as UAS traffic management.
%Hence, the proliferation of location-based services boosts the demand for a \emph{positioning system for complex multipath environments.}

For environments where line of sight (LOS) is dominant and multipath is scarce, various localization techniques have been proposed in the literature, majority of which are model-based. In these environments, physical models can describe the scattering surrounding quite well. These techniques mainly employ received signal strength indicator (RSSI), time of arrival (ToA), time difference of arrival (TDoA) and angle of arrival (AOA) measurements to pursue localization \cite{zane2020performance,sadowski2018rssi,ma2011toa,kaune2011accuracy,hejazi2020tensor,hejazikookamari2018novel,hejazi2013new,hejazi2014sar,hejazi2013lower,kookamari2017using}. One major issue with model-based techniques is that they normally require measurements of those parameters from multiple anchor nodes which may not be available in every environment.  %Moreover, model-based techniques fail to perform with acceptable accuracy in complex environments due to the scarcity of LOS and prevalence of multipath.% It may not be an exaggeration if someone claims that multipath is the most potent challenger of model-based localization techniques \cite{sen2013avoiding}. 
To address shortfalls of model-based techniques and to tackle localization for complex multipath environments, several data driven approaches have been proposed. Data-driven localization techniques are mainly called \emph{fingerprinting-based localization} \cite{zhang2019deep}. This type of localization generally constitutes gathering a dataset of a geo-tagged communication parameter (e.g. RSSI or channel state information (CSI)) all over the environment and training a neural network based on the dataset for online localization. Unfortunately, the data-driven approach has also failed to solve the problem of localization in complex multipath environments thus far. %, as they only perform well when the environment is completely static, while complex environments are inherently dynamic.
Data-driven approaches are highly dependent of an exorbitant campaign of gathering a geo-tagged dataset. Moreover, they need several hours of training. During these two relatively prolonged procedures, it is probable that the environment changes and the dataset becomes invalid \cite{zafari2019survey}. Some efforts have been conducted to taper the required dataset and reduce the training time \cite{he2019model}; however, in the best-case scenario they could reduce the initial required time to a couple of hours while these environments can change in a couple of seconds. Furthermore, it is almost impossible to gather data for every possible dynamic scenario and retrain the network. %Someone may think of crowdsourcing to relax the data gathering campaign, alas, the ground truth location on user devices are not available to tag the data, after all. 
%To the best of authors' knowledge, this work is the first attempt to tackle localization in complex dynamic environments with a data-driven perspective. 

Massive multi input multi output (MIMO) is a technique in wireless networks that utilizes numerous antennas mainly at base stations to take advantage of multipath effect to spatially multiplex users \cite{larsson2014massive}. %For the first time in the history of communication technology, MIMO could transform multipath from a foe to a friend \cite{witrisal2016high}.
 Massive MIMO is considered a core technology behind the revolution in ultra-high speed communications promised by 5G cellular networks \cite{jungnickel2014role}. To enable spatial multiplexing, base stations must identify the propagation environment from their antennas to the users’ antennas. This task is routinely conducted by measuring CSI. When measured perfectly, CSI preserves all information about scattering, fading, delays and power decay of the channel. Due to the rich information contained in CSI, it is considered an integral parameter for single-site fingerprinting-based localization. Vieira et. al. considered using a deep convolutional neural network (DCNN) trained by angle delay profiles (ADPs) for localization for the first time in the literature \cite{vieira2017deep}. Viera shows that in addition to memorizing the dataset, the trained DCNN can generalize localization to unknown location within the environment. Sun et. Al. trained two different DCNNs to pursue the localization task. The first network is similar to the regression network proposed by Viera. The second network incorporates two different blocks, the first block is a classification DCNN that defines to which grid cell the user location belongs, and the second block uses a weighted K-nearest neighbor (WKNN) algorithm to find the user location within the cell precisely \cite{sun2019fingerprint}. In \cite{ferrand2020dnn}, the authors design input features to make them robust to CSI impairments. They consider an autocorrelated version of CSI as the input to the CNN. %De Bast et. al.  consider uniform linear array (ULA) and uniform rectangular array (URA) at the base station. They show that transfer learning can radically reduce the time and number of samples required to train the network \cite{de2020csi}. 
 In \cite{de2020mamimo}, De Bast et. al. showed that CNN trained using CSI performs with centimeter accuracy in a static indoor environment. However, when a person is walking in the room the error increased by ten folds or more. To the best of our knowledge, this is the only work that examines the effect of dynamic scenarios on the accuracy of localization via a CNN trained using CSI. Unfortunately, the available literature mostly considers static scenarios and ignores the dynamic nature of complex scattering environments.
To address this shortcoming, in our work, we mainly focus on addressing localization in \emph{dynamic scenarios} leveraging a data-driven approach. To the best of authors' knowledge, this work is the first attempt to tackle localization in complex dynamic environments with a data-driven perspective.

Deep learning has shown outstanding performance in the field of computer vision (CV) so far. Among all various topics in the CV context, video surveillance, video frame prediction, video foreground and anomaly detection tackle highly dynamic problems \cite{nawaratne2019spatiotemporal,choi2019deep,sultani2018real}. In our work, we adopt some ideas from frame prediction literature to address data-driven localization in dynamic environments.  First, we prove that %ADPs are highly correlated in spatial domain when the environment is totally static. Then we conclude that 
time-series of ADPs preserves users’ movements assuming a static environment (the static environment is referred to as background (BG)). Consequently, it leads us to infer predictability of the next frame on an ADP temporal sequence based on previous frames using a predictive recurrent neural network. We model changes in the environment by LOS blockage, none LOS (NLOS) blockage, and NLOS addition, referred as foreground (FG). We propose an algorithm to discriminate between those ADPs which are accurate and those distorted by FG. Consequently, we propose \emph{DyLoc} to recover distorted ADPs and \emph{to estimate} the user location incorporating a WKNN block with a predictive recurrent neural network (PredRNN) block. To examine the performance of DyLoc we consider an indoor and an outdoor environments utilizing DeepMIMO dataset \cite{alkhateeb2019deepmimo}. We show that a trained CNN fails to estimate user location with acceptable accuracy in dynamic environments while the proposed technique can estimate user location with a decent accuracy. The main contributions of our work can be encapsulated as follow:
\begin{itemize}
    %\item Demonstrating ADP is highly correlated in spatial domain when the environment is completely static (BG)
    \item Proving that a time-series of ADPs can preserve users’ movements
    \item Showing ADP can be predicted based on previous frames of the ADP time-series
    \item Modeling changes in the environment as LOS blockage, NLOS blockage, and NLOS addition 
    \item Proposing DyLoc, a novel localization algorithm, that includes two steps: (i) an algorithm to detect distorted ADPs, and (ii) an algorithm to recover distorted ADPs and to estimate users’ location utilizing PredRNN and WKNN techniques. 
\end{itemize}
The rest of the paper is organized as follows. In Section \ref{prob}, we present the considered system and channel model and define ADP. We introduce motion preservation property of ADP in Section \ref{ADPProp}. In Section \ref{FG}, we express how we can model a dynamic propagation environment, where propagation paths can be blocked or added at anytime. We present a new prospective toward fingerprint gathering campaigns in Section \ref{rethink}. In Section \ref{DyLocAlg}, we introduce DyLoc to tackle the localization task in complex environments. In Section \ref{simsec}, we examine the performance of the proposed technique via various simulations. Finally, we conclude the paper in Section \ref{conc}.

\section{System and Channel Model}
\label{prob}
Assume we require to localize a single user, utilizing a single base station (BS) of a typical MIMO-Orthogonal frequency-division multiplexing (OFDM) wireless network. For the ease of exposition and similar to \cite{ali2017millimeter}, we suppose that the BS is equipped with a uniform linear array (ULA), with half wavelength spacing between two adjacent antennas, and a user's device has a single omni-directional antenna. The BS has $N_t$ antennas, and uses OFDM signaling with $N_c$ sub-carriers. We assume a geometric channel model between the BS and the user with $C$ distinguishable clusters. Moreover, each cluster constitutes $R_C$ distinguishable paths% between the BS and the user
. Each path can be characterized by a delay $\tau_{m}^{(k)}, k \in \{ 1, \dots,C\}, m \in \{ 1, \dots,R_C\}$, an AOA to the BS's antenna $\theta_{m}^{(k)}$ and a complex gain $\alpha_{m}^{(k)}$ \cite{ali2017millimeter}. Assuming a wide-band OFDM system, $\tau_{m}^{(k)} = n_{m}^{(k)} T_s$, where $T_s$ and $n_{m}^{(k)}$ denote the sampling duration and the sampled delay belonging to the path $m$ of the cluster $k$, respectively \cite{sun2019fingerprint}. Assuming these parameters, channel frequency response (CFR) for each sub-carrier $l$ can be written as \cite{alkhateeb2016frequency} 
\begin{equation}
    \boldsymbol{h}[l] = \sum_{k=1}^{C} \sum_{m=1}^{R_C} \alpha_{m}^{(k)} \boldsymbol{e}(\theta_{m}^{(k)}) e^{-j 2\pi \frac{l \: n_{m}^{(k)}}{N_c} } \,,
    \label{CSIdef}
\end{equation}
where $j$ denotes the imaginary unit and $\boldsymbol{e}(\theta)$ denotes the array response vector of the ULA given by
\begin{equation}
    \boldsymbol{e}(\theta) = [1,e^{-j2 \pi \frac{d cos(\theta)}{\lambda}},\dots,e^{-j2 \pi \frac{(N_t - 1)d cos(\theta)}{\lambda}}]^T\,,
\end{equation}
where $d$ is the gap between two adjacent antennas and $\lambda$ is the wavelength. Thus, the overall CFR matrix of the channel between the BS and the user can be expressed as
\begin{equation}
    \boldsymbol{H} = [\boldsymbol{h}[1],\dots,\boldsymbol{h}[N_c]]\,.
\end{equation}
This matrix commonly is referred to as CSI in the literature.

\begin{figure}[htbp]
 \vspace{-4mm}
    \centering
    \includegraphics[width=3.2in,height=2.2 in]{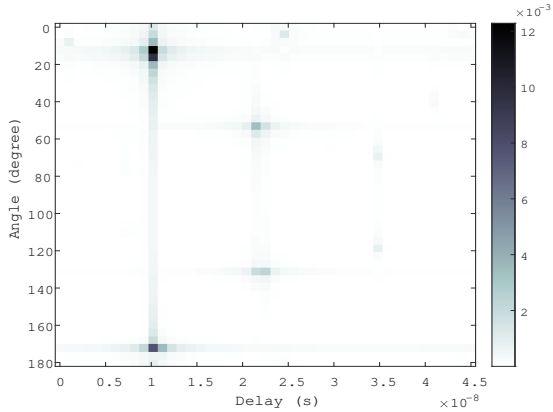}
    \caption {\small {A sample ADP image. Each pixel represents the absolute gain of the path with the corresponding AOA and the delay. Each "+"-like shape in the image shows a path cluster between the BS and the user.}}
    \vspace{-3mm}
    \label{testADP}
\end{figure}

%\subsection{Angle Delay profile (ADP)}
\emph{Angle Delay profile (ADP)} is a linear transformation of the CSI computed by multiplying it with two discrete Fourier transform (DFT) matrices. Let us define the DFT matrix $\boldsymbol{V} \in \mathbb{C}^{N_t \times N_t}$ as $$[\boldsymbol{V}]_{\: z,q} \overset{\Delta}{=} \frac{1}{\sqrt{N_t}} e^{-j 2 \pi \frac{\left(z(q-\frac{N_t}{2})\right)}{N_t} },$$ and $\boldsymbol{F} \in \mathbb{C}^{N_c \times N_c}$ as $$[\boldsymbol{F}]_{\: z,q} \overset{\Delta}{=} \frac{1}{\sqrt{N_c}} e^{-j 2 \pi \frac{zq}{N_c} }.$$Then ADP matrix $\boldsymbol{G}$ is defined as %follow 
\cite{sun2019fingerprint}
\begin{equation}
   \boldsymbol{G} = \boldsymbol{V}^H \boldsymbol{H} \boldsymbol{F}\,.
   \label{CSI2ADP}
\end{equation}
Now, let us define $[\boldsymbol{A}]_{\: z,q} = |\boldsymbol{G}_{z,q}|$, where $|.|$ denotes absolute value. Throughout this paper, we refer to $\boldsymbol{A}$ as ADP. When measured perfectly, CSI is a very rich data and preserves all scattering characteristics of the channel. However, when depicted in its raw format it is completely meaningless. On the other hand, referring to \cite{sun2018single}, the $(z,q)$ element of the ADP represents the absolute gain of $z^{th}$ AOA and $q^{th}$ delay as illustrated in Fig. \ref{testADP}. Therefore, we can simply make sense of ADP as a visual representation of all distinguishable paths between the user and the BS. For example, we can deduct from Fig.~\ref{testADP} that there is a LOS path cluster with AOA around $18^o$ and approximately $10^{-8}$s delay, and there are eight NLOS clusters between the user and the BS. Using ADP, we can cast the localization problem as a pattern recognition problem and take advantage of the rich literature of deep learning applications in CV \cite{sun2019fingerprint}. %(\hl{add two figure of ADP, indoor and outdoor})

%
%\vspace{-3mm}
\section{Scattering Environment}
%As we have discussed, CSI implicitly preserves all delay and AOA information belonging to all distinguishable paths between the user and the BS. Interestingly, ADP can reveal this information via a meaningful image. 
In this section, we discuss static and dynamic environments and how a user's motion reflects on ADP. We will show that when a user moves in a continuous track in the location domain, all paths in ADP also move in a continuous track, assuming a totally static environment. Moreover, we discuss dynamic changes in the environment and how we can model their effects on scatterings.
\subsection{Static Environment}
\label{ADPProp}
First, let us define what we mean by a static environment.
\emph{Definition 1.} \textbf{Static environment} is an environment including a user and at least one base station, in which nothing other than the user can move and all the materials of the surfaces remain the same within the environment. In this environment, scattering (influenced by propagation paths, decays, and delays%all propagation paths, decays, and delays
) remains unchanged. Moreover, the user's motion does not affect the scattering and does not block or add any path, while it may change visible paths between the BS and the user.

\emph{Theorem 1.} Let us consider a static environment where the user's movement does not change the paths between the BS and the user. Given a static environment, assume the user movement does not change the visible paths between the BS and the user. If user's position shifts by some very small positive amount $\delta d$, changes in delay of paths are limited to the following bounds
\begin{equation}
    \delta \tau_{m}^{(k)} \le \frac{\delta d}{v_c};\:\:\:\:\: 
    \forall{k} \in \{ 1, \dots,C\}, \forall{m} \in \{ 1, \dots,R_C\} \,,
\end{equation}
where $v_c$ denotes the speed of light and $\delta \tau_{m}^{(k)}$ denotes changes in the delay of the path $m$ of cluster $k$. Further, the following bound on the path AOA shift holds for any $\alpha > 1$
\begin{equation}
    \lim_{\delta d \to 0} \delta\theta_{m}^{(k)} \le \alpha\frac{\delta d}{d_{m}^{(k)}}; \:\:\:\:\: 
    \forall{k} \in \{ 1, \dots,C\}, \forall{m} \in \{ 1, \dots,R_C\} \,,
    \label{alpa}
\end{equation}

where  $\delta \theta_{m}^{(k)}$ and $d_{m}^{(k)}$
are changes in the AOA and the length of the path $m$ of cluster $k$, respectively.

\textbf{Proof.} When the user's position changes by $\delta d$, the length of each path from the BS to the user (LOS and NLOS) will change the same or less, thus the change in the delay is limited to $\frac{\delta d}{v_c}$.

\begin{figure}[ht]
    \vspace{-4mm}
    \centering
\includegraphics[width=2.7in,height=1.5in]{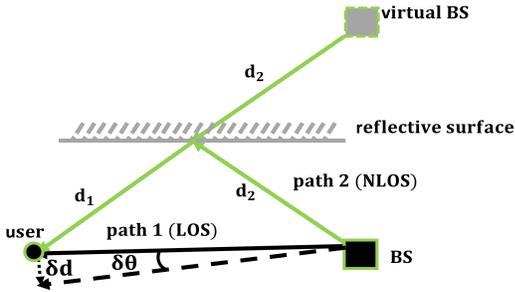}
    \caption {{LOS and NLOS paths geometry. The NLOS path can be considered as a LOS path from a virtual BS located at the reflection of the BS with respect to the reflective surface.}}
    \vspace{-2mm}
    \label{figgeo}
\end{figure}

Assuming LOS path (path 1 in Fig. \ref{figgeo}), change in the angle of the path is the maximum if the movement is perpendicular to the path. Thus, assuming the movement is perpendicular to the LOS, then $tan(\delta \theta) = \frac{\delta d}{d_{(path)}}$, where $d_{path}$ is the length of the path from the BS to the user and $\delta \theta$ is the change in AOA of signal from the user to the BS. Considering $\lim_{x \to 0} tan(x) \rightarrow x$, for any $\alpha > 1$ there is a $\delta d$ close enough to zero such that $\delta\theta_{m}^{(k)} \le \alpha\frac{\delta d}{d_{m}^{(k)}}$. 

Referring to \cite{shen2009use}, every NLOS path from the BS to the user, can be considered as a LOS path from a virtual BS to the user, with the same length (Fig. \ref{figgeo}). Thus, \eqref{alpa} holds for a NLOS path as well. $\square$    

Referring to Theorem 1, we can infer that, given a static environment and fixed paths between the BS and the user, any continuous user's movement in the location domain will result in a continuous movement in the angle-delay domain. In other words, when the user moves within the environment, all paths in the angle-delay domain start moving in a continuous track. Consequently, if we cascade consecutive ADPs to form a time-series data, the sequence is highly correlated temporally. 

Theorem 1 expresses that if a path exists in the ADP, it moves in a continuous track when the user moves. In addition to path motion, even in a static environment, it is possible that some paths dissipate and some new visible paths emerge during the user movement. Referring to \cite{sun2018single}, ADP is highly correlated in location domain and similarities between the ADPs decrease smoothly with respect to their physical distances. Thus, emergence and dissipation of paths occur very smoothly during the user movement.
%\NR{you have proved the simpler case with fixed paths, but refered to [30] for the harder and more general case without any proof. this can raise a question. also why do you say the emergence of and dissipation of paths are smooth?}.\FR{ In my view, the second problem is not a generalization of the first problem. These are 2 different problem: 1- if a path exists in the ADP, does it move in a continuous path? I dont think high spatial correlation between ADPs directly conclude this, so I proved it here. 2- How is path dissipation and emergence in the ADP? I believe high spatial correlation means that path dissipation/emergence wont happens abruptly. Suppose suddenly, 3 paths dissipate and 3 new paths emerge, ADPs are not highly correlated in spatial domain anymore. Moreover, in that work they introduce a metric for similarities between ADPs and they show that it is a decreasing and continuous function of physical distance, thus it is impossible that path emmergance/dissipation happens abruptly} 
Hence, the following conclusion can be inferred:

\textbf{Collorary 1:} Assuming a static environment, consecutive frames of time-series of the user measured ADPs show a continuous movement for all paths between the BS and the user.

This conclusion leads us to the idea that it may be possible to predict future ADP frames based on the time-series of past frames.

\subsection{Dynamic Behaviour of The Environment}
\label{FG}
Till now, we have discussed that a time-series of ADPs belonging to a certain user are highly correlated temporally, assuming a static environment. Nonetheless, complex scattering environments are normally highly dynamic. This means that several objects can move into and out of them and can change the scattering environment quickly and thoroughly. Dynamic changes in the environment result in the following changes in the static scattering environment:
\begin{enumerate}
    \item \textbf{LOS blockage}: a new object blocks the LOS path between the user and the BS.
    \item \textbf{NLOS blockage}: a new object blocks some NLOS paths between the user and the BS. 
    \item \textbf{NLOS addition}: scattering from surfaces of a new object adds some NLOS paths between the user and the BS.
\end{enumerate}
We assume that all objects and surfaces belonging to the static environments stay fixed and unchanged, therefore a LOS path already blocked by the static environment cannot get unblocked by the dynamic movements inside the environment.
%(Fig. \ref{figcity})
%\vspace{-2mm}
%\begin{figure}[htbp]
 %   \centering
  %  \includegraphics[width=3.5in,height=2.5 in]{Figcity.png}
   % \caption { Movement and changes in the environment may results in LOS blockage (blue line), NLOS blockage (red line) or NLOS addtion (green line).}
    %\vspace{-6mm}
    %\label{figcity}
%\end{figure}

\section{Rethinking of Fingerprint Gathering Campaigns}
\label{rethink}

To form a dataset of geo-tagged CSIs (or any other communication parameter), previous studies have generally considered measuring CSI at several locations inside a static environment. Such campaigns may be ineffective for user localization in dynamic scenarios in the first place, since few movements inside the environment may invalidate the whole dataset quickly. In this work, we introduce a new perspective toward these data gathering campaigns that can picture them as an integral part of any data-driven dynamic localization framework. In fact, by measuring CSI around a static environment, we map spatial distribution of all propagation paths within the static environment. It is like recording video footage from the whole static environment from the BS point of view. Such a footage can reveal all LOS and NLOS paths from any point inside the environment to the BS (or vice versa) in visible light band. Similarly, such measurement campaigns in radio frequency bands help us to understand the underlying scattering environment thoroughly, from the BS point of view. In the video processing literature, the underlying static environment and the changing environment are called "background" and "foreground", respectively. Here, we mimic the same pattern. In fact, measurement campaigns are quite obliging for understating the static scattering environment. Once the measurements are complete, we can utilize our understanding of BG to detect FG and employ proper algorithms to track changes and recover the true BG. Such a prospective toward fingerprinting, in conjunction with meaningful representation of the environment via ADP images, enables us to cast dynamic fingerprinting problem (or any other wireless communication problem that should deal with a dynamic scattering environment) as a video processing problem. Eventually, this redefinition enables us to take advantage of the rich literature of video surveillance, video frame prediction, and video foreground and anomaly detection in the context of computer vision to tackle wireless communication problems in dynamic environments.          
\section{Dyloc: Dynamic Localization Via Fingerprinting}
\label{DyLocAlg}
In this section, we introduce our proposed localization framework for dynamic environments. Suppose a user is moving inside an environment where there is a BS utilizing massive MIMO technology for communication as defined in Section \ref{prob}. %The BS continuously measures CSI between its' antennas and the user device with time interval $T_s$ between two consecutive measurements. 
Some of the measured CSIs may get affected by FG and get distorted compared to those of a static environment as explained in Section \ref{FG}. At first we train an off-the-shelf DCNN as introduced in \cite{sun2019fingerprint,vieira2017deep} to conduct localization assuming the environment is static. In this regard, we suppose we have a dataset of geo-tagged CSIs that maps the underlying BG exhaustively. To obtain a dataset of geo-tagged ADPs, we transform all CSIs to ADPs using \eqref{CSI2ADP}. We denote the dataset by $\boldsymbol{\Upsilon}$ which consists of ADPs paired by locations. Then we take the DCNN and train it based on the dataset to conduct the localization task when ADPs are not distorted. %Throughout this work we refer to the trained network as DCNN. This Network conducts the localization task whenever the measured CSI is not distorted. 

Now, we assume a stream of CSIs are measured consecutively to establish and maintain the link between the user and the BS. $\boldsymbol{H}_t$  denotes the CSI measured at time $t$, and $\boldsymbol{A}_t$ denotes the corresponding ADP. 
%obtained by applying linear transformation of \eqref{CSI2ADP} on the CSI.
 Primarily, we develop an algorithm to determine whether the measured ADP is distorted or not. In this regard, we pass $\boldsymbol{A}_t$ through DCNN and estimate the user location. Then we search the dataset and find nearby locations to the estimated location and compare the paired ADPs with the measured ADP, to see if there is at least one ADP among them that is similar to the measured one. To quantitatively measure similarity between two ADPs, we need a similarity metric. In \cite{sun2018single}, authors introduce the joint angle-delay similarity coefficient (JDASC) and prove that it is a decreasing function of physical distance. We observed that simple normalized correlation between two ADPs does the job as well. Thus, we define normalized correlation $\mathscr{S}$ as
\vspace{-1mm}
\begin{equation}
   \mathscr{S}(\boldsymbol{A},\hat{\boldsymbol{A}}) = \frac{\mathrm{vec}(\boldsymbol{A}).\mathrm{vec}(\hat{\boldsymbol{A}})}{||\boldsymbol{A}||_F||\hat{\boldsymbol{A}}||_F}, 
   \label{sim}
\end{equation} 

where $\boldsymbol{A},\hat{\boldsymbol{A}}$ denote two arbitrary ADPs, vec($.$) denotes an operator that concatenates columns of a matrix into a vector, operation $.$ denotes inner product and $||.||_F$ denotes Frobenius norm. %(In section(sim) we will show that this metric is also decreasing with physical distance). 
If there is at least one ADP from the neighboring locations whose similarity to the measured ADP is more than a predefined threshold ($\mathrm{thr}_2$), we label the ADP as "accurate", otherwise we label it as "distorted". The algorithm for distorted ADP detection is summarized in Algorithm \ref{alg1}. The rationale behind the proposed algorithm stems from our discussion in Section \ref{ADPProp} that ADP is highly correlated in the location domain. Since the DCNN is solely trained by accurate ADPs, it will return inaccurate locations when faced with distorted ADPs. Hence, nearby ADPs will not show high similarities with the measured ADP. On the other hand, if the measured ADP is accurate, there will be an ADP in the nearby ADPs that looks very similar to the measured one.

\begin{algorithm}
\algsetup{
linenosize=\scriptsize,
linenodelimiter=:
}
\caption{Distorted ADP Detection}
\begin{algorithmic}[1]

\label{alg1}
\footnotesize
\REQUIRE{measured CSI at time $t$, $\boldsymbol{H}_t$}; $\mathrm{thr}_1,\mathrm{thr}_2$ \\
\ENSURE{$\boldsymbol{H}_t$ is distorted or accurate }
\vspace{+1mm}
\STATE{Convert $\boldsymbol{H}_t$ to ADP $\boldsymbol{A}_t$ using \eqref{CSI2ADP}}  
\vspace{2pt}
\STATE{Apply $\boldsymbol{A}_t$ to DCNN and get the location estimation $\boldsymbol{x}_t$}  
\vspace{2pt}
\STATE{$\mathbb{X}_t$ $\leftarrow$ find all paired ADPs in the dataset where their tagged locations are closer than $\mathrm{thr}_1$} to $\boldsymbol{x}_t$ 
\STATE \emph{flag} $\leftarrow$ 0
\FOR{all ADPs $\boldsymbol{A} \in \mathbb{X}_t$}
    \STATE{$s$ $\leftarrow$ $\mathscr{S}\left(\boldsymbol{A},\boldsymbol{A}_t\right)$ }
        \IF {$s > \mathrm{thr}_2$}
            \STATE{\emph{flag} $\leftarrow$ 1}
        \ENDIF
\ENDFOR
\IF{\emph{flag} == 1} 
    \RETURN {accurate}
\ELSE
    \RETURN {distorted}
\ENDIF\\
    \end{algorithmic}
\end{algorithm}
\vspace{0mm}

If the measured ADP is accurate, we can simply use DCNN for localization. On the other hand, if it is distorted, we propose to use WKNN along with a video frame prediction algorithm to recover the true ADP and conduct localization. Based on our discussion in Sections \ref{ADPProp} and \ref{rethink}, time-series of accurate ADPs are highly correlated temporally. Thus, given time-series of accurate ADPs before facing a distorted ADP, we try to predict the next frame of the time-series using a predictive recurrent neural network (PredRNN). PredRNN is a frame prediction algorithm that tries to learn dependencies between consecutive frames and uses this knowledge to predict the next frame of the sequence. Frame prediction is a challenging problem in CV and several algorithms have been published to address it \cite{liang2017dual,liu2018future}. We chose PredRNN mainly because it shows promising results on the radar echo dataset. The detailed discussion about PredRNN can be found in \cite{wang2017predrnn}. In this work, we assume that we have a sequence of past accurate ADPs with length $f \in \mathbb{N}$ at time $t$, $\boldsymbol{A}_{t-fT_s},\dots,\boldsymbol{A}_{t-T_s}$ denoted by $\mathcal{A}_t$. We train the network based on a dataset of random walks which we generate using $\Upsilon$. In Section \ref{simset}, we will describe how we generate the moving dataset, the PredRNN structure and how we train it, in details.

\begin{figure*}
    \centering
    \includegraphics[width=6.5in,height=2 in]{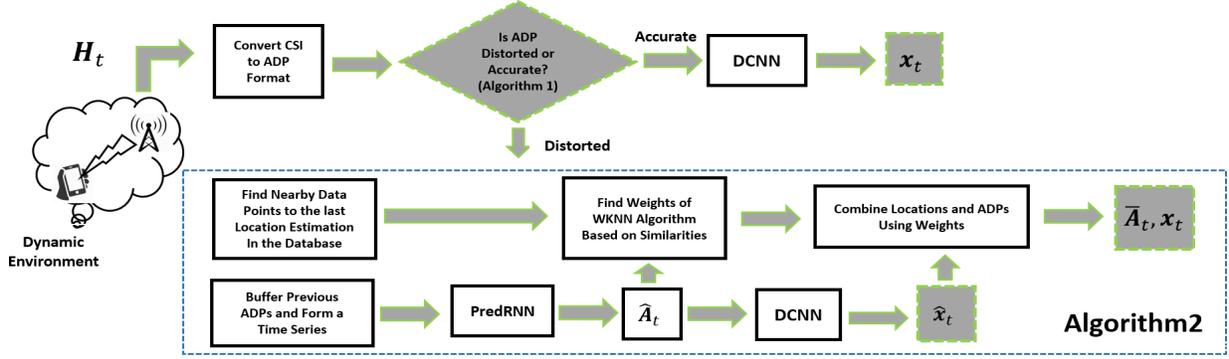}
    \caption {\small{ Block Diagram of the proposed localization framework DyLoc.}}
    \vspace{-1mm}
    \label{figblock}
\end{figure*}

After detecting a distorted ADP, we pass $\mathcal{A}_t$ through the trained PredRNN to predict the accurate ADP denoted by $\hat{\boldsymbol{A}_t}$. Next, we pass $\hat{\boldsymbol{A}_t}$ through DCNN to obtain an initial estimation of the user location $\hat{\boldsymbol{x}}_t$. In addition to frame prediction, in light of the fact that ADPs are highly correlated in location domain, we can take the last location estimation based on the last accurate ADP and find nearby locations in the database and use them to reach a better location estimation and recover the true ADP. To clarify, if some paths in the ADP get blocked or added, the remaining paths pose correlation with nearby ADPs. Thus, using the similarity criteria \eqref{sim}, we can extract the residual similarities between the distorted ADP and nearby ADPs. Moreover, we calculate the similarity between the distorted ADP and the predicted one. Now we are able to combine nearby locations and ADPs and the estimated location and ADP via a WKNN algorithm to obtain a better location estimation and  recover ADP. Weights of WKNN can be determined directly from calculated similarities. Hence, the location and the true ADP can be estimated as 
\vspace{0mm}
\begin{equation}
    \boldsymbol{x}_t = \sum_{x \in \mathscr{N}} w_x \boldsymbol{x};\:\:\:\:\: \bar{\boldsymbol{A}_t} = \sum_{x \in \mathscr{N}} w_x \boldsymbol{A}_x\,,
\end{equation}
\vspace{-3mm}

where $\boldsymbol{x}_t$ denotes the estimated location and $\bar{\boldsymbol{A}_t}$ denotes the recovered ADP, $\boldsymbol{A}_x$ is ADP at location $\boldsymbol{x}$, $\mathscr{N}$ denotes union of the set of nearby locations added by the predicted location, and $w_x$ is the weight, given by
\vspace{-0mm}
\begin{equation}
    w_x = \frac{\mathscr{S}(\boldsymbol{A}_t,\boldsymbol{A}_x)}{\sum_{\boldsymbol{A} \in \mathscr{A}}\mathscr{S}(\boldsymbol{A}_t,\boldsymbol{A})}\,,
\end{equation}
\vspace{-2mm}

where $\mathscr{A}$ denotes the set of ADPs corresponding to locations in $\mathscr{N}$. Finally, we can estimate the user location at time $t$ (denoted by $\boldsymbol{x}_t$) and recover the ADP ($\bar{\boldsymbol{A}_t}$) which can be used for future location estimations. Algorithm \ref{alg2} summarizes the proposed algorithm for ADP recovery and location estimation. Moreover, Fig. \ref{figblock} summarizes the end-to-end DyLoc localization framework.

\begin{algorithm}
\algsetup{
linenosize=\scriptsize,
linenodelimiter=:
}

\caption{Location Estimation When ADP Is Distorted}
\begin{algorithmic}[1]
\label{alg2}

\footnotesize
\REQUIRE{time-series of accurate ADPs $\mathcal{A}_t:\boldsymbol{A}_{t-T_sf},\dots,\boldsymbol{A}_{t-T_s}$\\
\:\:\:\: distorted ADP $\boldsymbol{A}_{t}$\\
\:\:\:\: geo-tagged ADP dataset $\Upsilon$\\
\:\:\:\: the previous estimated location $\boldsymbol{x}_{t-T_s}$\\
\:\:\:\: $\mathrm{thr}_3$}
\ENSURE{The recovered ADP $\bar{\boldsymbol{A}_t}$, and location estimation $\boldsymbol{x}_t$}
\vspace{+1mm}
\STATE{\textbf{Frame prediction}}  
\vspace{2pt}
\STATE{Apply $\mathcal{A}_t$ to PredRNN and get the predicted ADP   $\hat{\boldsymbol{A}_t}$}  
\vspace{2pt}
\STATE{Apply $\hat{\boldsymbol{A}_t}$ to DCNN and get the location estimation $\hat{\boldsymbol{x}}_t$}
\STATE{\textbf{WKNN}}
%\STATE{Apply $\boldsymbol{A}_{t-T_s}$ to DCNN and get the previous credible location estimation $\boldsymbol{x}_{t-T_s}$}
\STATE{$\mathscr{N}$ $\leftarrow$ find all data points in the database which their locations are closer than $\mathrm{thr}_3$ to $\boldsymbol{x}_{t-T_s}$}
\STATE{$\mathscr{N}_A$ $\leftarrow$ paired ADPs of locations in $\mathscr{N}$}
\STATE{$\mathscr{N}$ $\leftarrow$ $\mathscr{N} \cup \hat{\boldsymbol{x}}_t$}
\STATE{$\mathscr{N}_A$ $\leftarrow$ $\mathscr{N}_A \cup \hat{\boldsymbol{A}_t}$}
\STATE{$\boldsymbol{x}_t \leftarrow \sum_{x \in \mathscr{N}} w_x \boldsymbol{x}_x; w_x = \frac{\mathscr{S}(\boldsymbol{A}_t,\boldsymbol{A}_x)}{\sum_{\boldsymbol{A} \in \mathscr{N}_A}\mathscr{S}(\boldsymbol{A}_t,\boldsymbol{A})}$}
\STATE{$\bar{\boldsymbol{A}_t} \leftarrow \sum_{x \in \mathscr{N}} w_x \boldsymbol{A}_x$}
\RETURN{$\bar{\boldsymbol{A}_t}$,$\boldsymbol{x}_t$}
\end{algorithmic}
\end{algorithm}
\vspace{-1mm}
\section{Simulations}
\label{simsec}
In this section, the performance of our proposed localization framework for an indoor and an outdoor environments is studied. In Section \ref{dataset}, we present the dataset that we used for static environment fingerprinting. We define the structure of the DCNN and how to train it in Section \ref{DCNNset}. Then, we clarify how we generate the moving dataset using the static dataset in Section \ref{MD}. Further, we explain the structure of the PredRNN and how we trained it. Next, we express dynamic scenarios generated to test the performance of DyLoc. Finally, in Section  \ref{ResDis} we dive deep into evaluating the performance of DyLoc for various dynamic scenarios and compare it with the state-of-the-art. \footnote{The authors release their codes in the following link
”https://github.com/FarzamHejaziK/DyLoc”.}
%\subsection{Static Dataset and DCNN Setup}
%\label{dataset}
%In this section, the performance of the proposed localization framework for an indoor and an outdoor environments is studied. In section \ref{dataset} we present the dataset we used for static environment fingerprinting. Moreover, we define the structure of the DCNN and how to train it. Then we clarify how we generate the moving dataset using the static dataset in section \ref{MD}. Further, we explain the structure of the PredRNN and how we train it. Next, we express dynamic scenarios that we have generated to test the performance of DyLoc. Finally, in section \ref{ResDis} we deep dive into evaluating the performance of DyLoc for various dynamic scenarios and compare it with the state-of-the-art.
\vspace{-1mm}
\subsection{Static Datasets}
\label{dataset}

In this work, we use DeepMIMO dataset to generate CSI datasets in static environments\footnote{http://www.deepmimo.net/}. % DeepMIMO is a generic millimeter wave and MIMO dataset first introduced in 2019 \footnote{http://www.deepmimo.net/}. DeepMIMO utilizes ray-tracing data obtained fromRemcom Wireless InSite \cite{RemComm} to generate its generic dataset.
Thus far, DeepMIMO has introduced for one outdoor and two indoor environments. We picked one indoor and one outdoor environments for our simulations.%\KV{propose deleting this sentence: The details about each environment will follow.}%is explained in the following two Sections.

\subsubsection{Outdoor Environment}
\label{MD} To generate an outdoor environment we select DeepMIMO outdoor scenario number 1 (O1) at $3.5$ GHz band. "O1" is an urban environment of two streets and one intersection. We suppose only BS number 2 (BS2) is working and it has been equipped with a ULA with $N_t=$ 64 antennas aligned with y-axis. We set the OFDM bandwidth to 10 MHz and $N_c$ = 64. We also set the number of paths to 25. Furthermore, we only generate a dataset for R1 to R1100 (Rows 1 to 1100, show locations of data points in DeepMIMO) , therefore the datset constitutes of 199100 data points. Table \ref{tb1} summarizes the dataset parameters. 

\subsubsection{Indoor Environment}
\label{I3set}
We picked DeepMIMO indoor scenario number 3 "I3" at $60$ GHz to emulate an indoor environment. "I3" simulates a $10 m\times11 m$ conference room and its hallway. We assume only access point number 2 (BS2) is working. Other parameters that set up the indoor propagation environment are reflected in Table \ref{tb1}. 
%since the number of training samples is only 66550 and the network can converge very fast.

\subsection{DCNN Setup}
\label{DCNNset}

As we explained in Section \ref{intro}, in \cite{sun2019fingerprint} two different DCNNs have been introduced to pursue localization using ADPs. We refer to the first setup that utilizes a regression network as DCNN and the second setup that uses a classification network along with WKNN as DCNN+WKNN. 
%The first one is a regression neural network that gets ADP and returns the estimated location when trained. The second one constitutes of two blocks. \textbf{The first block is a classification network that determines to which cell of a grid -that covers the whole area of interest- the location belongs.} The second block is a WKNN algorithm that estimates the exact location within the determined cell.
In our work, we train the DCNN to learn the background scattering environment as a part of DyLoc as described in Section \ref{DyLocAlg}. We also compare the performance of DCNN and DCNN+WKNN with DyLoc. %\KV{Recommend deleting: The paremeter setup for the two networks for each environment will follow.}
%\KV{somewhere in this section explain that one is a regression model and one is classification}

\begin{table}[htbp]
\vspace{-2mm}
 \begin{center}
\caption{\small{"O1" and "I3"  DeepMIMO datasets' parameters}}

\begin{tabular}{ |c|c|c| } 
 \hline
 \textbf{Parameter} &  \begin{tabular}{@{}c@{}}\textbf{Outdoor} \\ \textbf{Scenario (O1)} \end{tabular} & \begin{tabular}{@{}c@{}}\textbf{Indoor} \\ \textbf{Scenario (I3)} \end{tabular}\\
 
 \hline
 Frequency Band & 3.5 GHz  & 60 GHz\\ 
  \hline
  Bandwidth & 10 MHz  & 0.5 GHz\\ 
  \hline
 BS  & BS2 & BS2  \\ 
  \hline
 Antenna  & ULA & ULA \\ 
 \hline
 Antenna Elements ($N_t$) & 64 & 32 \\
 \hline
 Antenna Alignment & y-axis & x-axis \\
  \hline
 Sub-carrier Number ($N_c$) & 64 & 32  \\
  \hline
   Path Number & 25 & 25  \\
  \hline
 Locations & R1 to R1100  & R1 to R550  \\
\hline
\end{tabular}
\vspace{-3mm}
 \label{tb1}
\end{center}  
\end{table}

\subsubsection{Outdoor Environment} Based on the architecture presented in \cite{sun2019fingerprint}, we choose the parameters presented in Table \ref{tb2} for the DCNN setup. We use Max-pooling for pooling layers with size $2\times2$ and ReLU for activation function. We set training epochs to 500. The setup for the classification network in DCNN+WKNN is the same as DCNN while we add a Softmax layer to the network to conduct classification. Defining area of interest as the set of all locations in the dataset, we assume a $18\times55$ grid on the area of interest (18 equally-spaced segments in the x-direction and 55 segments in the y-direction). The classification network is trained to determine to which cell of the grid an input ADP belongs. Then using a WKNN technique with $k=3$ we estimate the location.

 \subsubsection{Indoor Environment} Parameters setup in Section \ref{I3set}  results in $32\times32$ ADP images in this scenario.  We train a 5-layer regression DCNN with parameters as in Table \ref{tb2} to learn the underlying propagation environment. The number of training epochs in this simulation is set to 200. The classification CNN setup in DCNN+WKNN technique is the same as the DCNN setup. Similar to the outdoor environment we assume a $18\times55$ grid % \KV{is this correct? I would expect a different grid size since the total area is smaller} (\FR{The size of the area is smaller, but the carrier frequency is much higher and bandwidth is larger(50 time larger), and the distance between the BS and user is much shorter, so it is not unexpected})
on the area of interest. We also set $k=3$ for the WKNN technique. 
 
 \begin{table}[htbp]
\vspace{-1mm}
 \begin{center}
\caption{\small{DCNN setup for "O1" and "I3"}}

\begin{tabular}{ |c|c|c|c|c| } 
 \hline
 \textbf{Layer} & \begin{tabular}{@{}c@{}}\textbf{Kernel} \\ \textbf{Size (O1)} \end{tabular}& \begin{tabular}{@{}c@{}}\textbf{Kernel} \\ \textbf{Number (O1)} \end{tabular} & \begin{tabular}{@{}c@{}}\textbf{Kernel} \\ \textbf{Size (I3)} \end{tabular} & \begin{tabular}{@{}c@{}}\textbf{Kernel} \\ \textbf{Number (I3)} \end{tabular} \\
 \hline
 1 & $32 \times 32 \times 1$ & 2 & $16 \times 16 \times 1$ & 4   \\ 
  \hline
 2 & $16 \times 16 \times 2$ & 4 & $8 \times 8 \times 4$ & 8 \\ 
  \hline
3 & $8 \times 8 \times 4$ & 8 & $7 \times 7 \times 8$ & 16 \\ 
 \hline
4 & $7 \times 7 \times 8$ & 16 & $5 \times 5 \times 16$ & 32 \\
 \hline
5 & $5 \times 5 \times 16$ & 32 & $3 \times 3 \times 32$ & 64 \\
  \hline
6 & $3 \times 3 \times 32$ & 64 &  &   \\

\hline
\end{tabular}
 \label{tb2}
 \vspace{-4mm}
\end{center}  
\end{table}
 \begin{table*}[h]
%\scriptsize
  \centering
  \caption{{ Location estimation RMSE in meter of the last 10 frames of the time-series employing DyLoc, DCNN \cite{sun2019fingerprint}, DCNN+WKNN \cite{sun2019fingerprint}, and PredRNN \cite{wang2017predrnn} for LOS blockage, NLOS blockage, NLOS addition scenarios at outdoor environment "O1" and outdoor environment "I3". PredRNN error shows location estimation error of the predictive arm of DyLoc.}}
  %\scriptsize
    \begin{tabular}{ccccccccccccc}
   \cmidrule{4-13}        
    &       &       &
\multicolumn{10}{c}{Frame Number} \\
\cmidrule{4-13}          &       &  RMSE(m)      & 11    & 12    & 13    & 14    & 15    & 16    & 17    & 18    & 19    & 20 \\
\hline
    \multirow{11}[8]{*}
    {\begin{sideways}
    {\fontsize{13}{15}\selectfont{Scenario O1}} 
    \end{sideways}} & \multirow{3}[1]{*}{LOS Blockage} & DyLoc   & \textbf{0.37} & \textbf{0.53} & \textbf{0.69} & \textbf{0.85} & \textbf{1.00} & \textbf{1.14} & \textbf{1.29} & \textbf{1.43} & \textbf{1.57} & \textbf{1.71} \\
          &       & DCNN   & 25.55 & 25.14 & 25.51 & 25.01 & 25.18 & 24.58 & 25.17 & 25.72 & 24.11 & 24.96 \\
          &       & DCNN + WKNN  & 181.66 & 179.30 & 178.88 & 178.90 & 175.11 & 178.90 & 183.42 & 176.93 & 180.52 & 176.00 \\
          &       & PredRNN   & 2.05  & 1.96  & 2.32  & 2.45  & 2.56  & 2.72  & 2.88  & 3.05  & 3.24  & 3.56 \\
\cmidrule{2-13}          & \multirow{3}[2]{*}{NLOS Blockage} & DyLoc   & \textbf{0.30} & \textbf{0.38} & \textbf{0.45} & \textbf{0.52} & \textbf{0.58} & \textbf{0.65} & \textbf{0.71} & \textbf{0.78} & \textbf{0.83} & \textbf{0.89} \\
          &       & DCNN   & 10.15 & 10.69 & 10.95 & 11.05 & 10.60 & 11.23 & 11.06 & 10.95 & 10.98 & 10.65 \\
           &       & DCNN + WKNN  & 36.17 & 34.66 & 34.52 & 34.45 & 38.95 & 35.32 & 36.86 & 36.23 & 35.72 & 39.38 \\
          &       & PredRNN   & 2.05  & 2.09  & 2.16  & 2.16  & 2.21  & 2.22  & 2.21  & 2.26  & 2.28  & 2.34 \\
\cmidrule{2-13}          & \multirow{3}[1]{*}{NLOS Addition} & DyLoc   & \textbf{0.30} & \textbf{0.42} & \textbf{0.53} & \textbf{0.63} & \textbf{0.72} & \textbf{0.82} & \textbf{0.91} & \textbf{0.99} & \textbf{1.08} & \textbf{1.16} \\
          &       & DCNN   & 15.09 & 15.09 & 15.26 & 15.44 & 15.00 & 15.21 & 14.80 & 15.36 & 15.12 & 15.66 \\
           &       & DCNN + WKNN  & 55.86 & 57.43 & 55.06 & 55.45 & 54.41 & 56.40 & 53.86 & 54.69 & 53.70 & 55.59 \\
          &       & PredRNN   & 2.05  & 2.23  & 2.39  & 2.38  & 2.38  & 2.40  & 2.44  & 2.48  & 2.51  & 2.57 \\
\hline
   \multirow{11}[8]{*}
    {\begin{sideways}
    {\fontsize{13}{15}\selectfont{Scenario I3}}
    \end{sideways}} & \multirow{3}[1]{*}{LOS Blockage} & DyLoc   & \textbf{0.04} & \textbf{0.04} & \textbf{0.04} & \textbf{0.05} & \textbf{0.05} & \textbf{0.06} & \textbf{0.06} & \textbf{0.07} & \textbf{0.07} & \textbf{0.08} \\
          &       & DCNN   & 0.93  & 0.93  & 0.93  & 0.92  & 0.93  & 0.92  & 0.93  & 0.93  & 0.93  & 0.93 \\
           &       & DCNN + WKNN  & 2.02 & 2.00 & 2.04 & 2.03 & 1.99 & 2.01 & 2.03 & 1.99 & 2.00 & 1.99 \\
          &       & PredRNN   & 0.15  & 0.16  & 0.16  & 0.16  & 0.17  & 0.17  & 0.17  & 0.17  & 0.17  & 0.17 \\
\cmidrule{2-13}          & \multirow{3}[2]{*}{NLOS Blockage} & DyLoc   & \textbf{0.05} & \textbf{0.05} & \textbf{0.06} & \textbf{0.06} & \textbf{0.07} & \textbf{0.07} & \textbf{0.07} & \textbf{0.08} & \textbf{0.08} & \textbf{0.08} \\
          &       & DCNN   & 0.47  & 0.47  & 0.47  & 0.48  & 0.48  & 0.48  & 0.47  & 0.46  & 0.46  & 0.47 \\
           &       & DCNN + WKNN  & 1.12 & 1.10 & 1.07 & 1.11 & 1.10 & 1.06 & 1.11 & 1.13 & 1.12 & 1.10 \\
          &       & PredRNN   & 0.15  & 0.16  & 0.16  & 0.17  & 0.17  & 0.18  & 0.18  & 0.18  & 0.18  & 0.18 \\
\cmidrule{2-13}          & \multirow{3}[2]{*}{NLOS Addition} & DyLoc   & \textbf{0.04} & \textbf{0.04} & \textbf{0.05} & \textbf{0.05} & \textbf{0.06} & \textbf{0.06} & \textbf{0.07} & \textbf{0.07} & \textbf{0.08} & \textbf{0.08} \\
          &       & DCNN   & 0.20  & 0.20  & 0.20  & 0.19  & 0.19  & 0.19  & 0.19  & 0.19  & 0.19  & 0.19 \\
          &       & DCNN + WKNN  & 1.35 & 1.27 & 1.33 & 1.39 & 1.32 & 1.36 & 1.34 & 1.37 & 1.33 & 1.32 \\
          &       & PredRNN   & 0.15  & 0.16  & 0.16  & 0.17  & 0.17  & 0.18  & 0.18  & 0.18  & 0.18  & 0.18 \\
    \bottomrule
    \end{tabular}%
  \label{bigres}%
  \vspace{-3mm}
\end{table*}%

%\begin{table}[htbp]
% \begin{center}
% \vspace{-4mm}
%\caption{{DCNN setup for "I3"}}
%
%\begin{tabular}{ |c|c|c| } 
 %\hline
 %&\textbf{Layer} & \textbf{Kernel Size} & \textbf{Kernel Number}\\

 %\hline
% 1 & $16 \times 16 \times 1$ & 4   \\ 
%  \hline
% 2 & $8 \times 8 \times 4$ & 8 \\ 
%  \hline
%3 & $7 \times 7 \times 8$ & 16  \\ 
% \hline
%4 & $5 \times 5 \times 16$ & 32  \\
% \hline
%5 & $3 \times 3 \times 32$ & 64  \\

%\hline
%\end{tabular}
% \label{tb3}
% \vspace{-10mm}
%\end{center}  
%\end{table}

\vspace{-1mm}
\subsection{Moving Dataset and PredRNN Setup}
\label{simset}

To train PredRNN and to emulate dynamic scenarios, we needed moving datasets. We generated moving datasets utilizing static datasets introduced in the previous section. Since the static datasets define the propagation environment thoroughly, they can be used to form dynamic datasets by stacking adjacent locations and paired ADPs. Both "O1" and "I3" datasets assume a grid on the corresponding environment  ("O1" assumes a 20 cm distance between two adjacent grid points and "I3" assumes grid of 1 cm apart). We take advantage of the underlying grid and form our moving dataset. To initialize each movement, we suppose the user is randomly placed on one of the grid points. Then at each step of the movement the user moves to one of the adjacent grid points. %with equal probability. %(Fig \ref{RW})%

 We assume the movement continues for $f$ consecutive steps. Next, we stacked all locations and paired ADPs together to form and save one sequence. We consider 2 modes for user random walks:
\begin{itemize}
\item \textbf{Mode 1:} User chooses its direction only at the first time-step and continues the same direction in the following time-steps. If the user reaches the boundaries of the environment, the user chooses another direction from possible remaining directions randomly. 
\item \textbf{Mode 2:} User goes for a complete random walk and at each time-step chooses a new direction.
\end{itemize}
To train PredRNN, we generated the whole dataset based on mode 1 movements since we expect that PredRNN is not able to predict ADPs stem from fully random walks. Thus, we generated 10000 sequences each with length 11. We employed the PredRNN presented in \cite{wang2017predrnn} with the same setup and parameters. We fed the first 10 ADP frames to the network and optimized the network to predict the $11^{th}$ ADP; so we do not use the location sequence for training, we only make use of the paired ADPs. Eventually, we expect the trained PredRNN is capable of predicting the next ADP in the sequence, given the 10 last accurate ADPs. 

\subsubsection{Dynamic Test Scenarios}
To evaluate the performance of the proposed DyLoc algorithm, we generate 1000 sequences of length 20 for each of the indoor and the outdoor environments. Half of these sequences is generated based on Mode 1 movements and half of them based on Mode 2. We assume the first 10 frames of each sequence consist of accurate ADPs and FG does not affect them. On the other hand, we assume the last 10 frames of each sequence are distorted based on one of the following scenarios:

\begin{itemize}
    \item \textbf{LOS Blockage}: we assume that the most powerful path between the BS and the user gets blocked, thus we eliminate this path from all 10 ADP frames.
    \item \textbf{NLOS Blockage}: we assume that the second most powerful path between the BS and the user gets blocked, thus we wipe out this path from all 10 ADP frames.
    \item \textbf{NLOS Addition}: we add a path 6dB weaker than the strongest path arbitrarily located in the ADP image to all 10 ADP frames.
    
\end{itemize}

We inspect DyLoc performance for the above 3 scenarios for both "O1" and "I3" environments and compare it with the state-of-the-art DCNN and DCNN+WKNN algorithm \cite{sun2019fingerprint}. 
%\begin{figure}[ht]
   % \centering
    %\includegraphics[width=3.7in,height=3 in]{fig3.pdf}
   % \caption {\small{ Forming random walks based on the location grids of static dataset. Blue dots shows the grid points, the green dot shows user's location on the grid and blue arrows show users' possible directions. At each user can select one of the adjacent grid point with equal probability and move there.}}
   % \label{RW}
%\end{figure}

 \subsection{Results and Discussion}
 \label{ResDis}
 In this section, we present the performance of DyLoc for the scenarios introduced in Section \ref{simset} for the outdoor environment "O1" and the indoor environment "I3".
 
 \subsubsection{Outdoor Environment}
 Table \ref{bigres} summarizes root mean square error (RMSE) of location estimation for the distorted frames for the 3 scenarios using DyLoc and compares it with DCNN and DCNN+WKNN. For the first 10 accurate frames, the RMSE for all scenarios using DCNN and DCNN+WKNN is 19 cm and 14.5 cm, respectively. Since DyLoc uses the same DCNN when frames are accurate, the DyLoc accuracy is 19 cm. However, when the frames are distorted, DCNN error proliferates to more than 20 m when LOS is blocked, more than 10 m when NLOS is added, and more than 7.5 m when NLOS is blocked. This huge surge in error is because DCNN has not been trained for localization based on distorted ADPs.  Moreover, DCNN error is the highest when LOS is blocked. Since the outdoor environment is not a rich scattering environment and generally there are 2-3 paths between the BS and the user, the LOS path contains the most valuable information in the ADP, so if it is blocked, the DCNN loses its most important clue for localization. When NLOS paths are distorted, the effect of NLOS addition is more than NLOS blockage since we assume a very strong path is added to the ADP. The DCNN+WKNN performance is even worse than DCNN facing distorted ADPs and the error is more than 180 m when LOS is blocked, more than 30 m when NLOS is added, and more than 50 m when NLOS is blocked. Since the classification network cannot find the correct cell that the distorted ADP belongs to, the WKNN layer does not perform well and the performance plunges drastically.
In contrast to DCNN and DCNN+WKNN, DyLoc performs with high accuracy when faces distorted ADPs. Regarding the TABLE \ref{bigres}, as frame number increases the DyLoc error increases. This is totally expected since the prediction error of the previous predicted frames propagates to the next frame and results in a higher error at the next frame. Nevertheless, the error remains less than 1.8 m, 1.2 m, and 0.9 m, for LOS blockage, NLOS addition, and NLOS blockage for all the distorted frames, respectively. %Moreover, in 90 percent of simulations the error is less than 3.1m, 2.5m and 2.2m for LOS blockage, NLOS addition and NLOS blockage for all the distorted frames, 
These error values show a quite promising performance by DyLoc in the outdoor environment for dynamic scenarios.

Referring to Table \ref{bigres}, RMSE of location estimation via the predictive arm of Algorithm \ref{alg2} (PredRNN) -before incorporating WKNN ($\hat{\boldsymbol{x}}_t$)- for all distorted frames and for the 3 scenarios are expressed. The error changes between 2.0 m to 2.4 m for the NLOS blockage scenario,  2.0 m to 2.6 m in the NLOS addition scenario, and 2.0 m to 3.6 m in the LOS blockage scenario. The error is increasing with a higher rate when LOS gets blocked since the measured ADP can help us the least to improve our prediction. Moreover, % we should mentioned that,
  in the LOS blockage scenario, especially when we have only one path between the BS and the user, and it gets blocked (i.e. the connection is lost),  $\hat{\boldsymbol{x}}_t$ is our exclusive source of location estimation. %Although pure prediction errors are higher than the final error of DyLoc,
  Thus, the predictive arm is very crucial for location estimation in the LOS blockage scenario. Additionally, when we incoroprate WKNN to form DyLoc, {the error values decrease to 0.3 m (frame 11) and 0.89 m (frame 20), 0.3 m (frame 11) and 1.16 m (frame 20), and 0.37 m (frame 11) and 1.71 m (frame 20)  for the NLOS Blockage, NLOS addition, and LOS blockage scenarios, respectively.} These results show that the WKNN arm is pretty successful in reducing the total estimation error. Consequently, both WKNN and PredRNN arms are crucial for accurate location estimation and robustness to FG dynamic changes.
 
 \subsubsection{Indoor Environment}
  
  Similar to the outdoor environment, we compare DCNN, DCNN+WKNN, and DyLoc algorithms in the indoor environment "I3" for the 3 dynamic scenarios. In contrast to the outdoor environment, the indoor environment is a rich scattering environment and there are several propagation paths between the BS and the user. The RMSE for accurate frames utilizing DCNN is 5 cm and DCNN+WKNN is 4.5 cm. Unlike the outdoor scenario, the indoor scenario DCNN performs much better when NLOS is added to the ADP. This happens because the scattering environment is rich and NLOS addition can be filtered out by DCNN.

  As Table \ref{bigres} expresses, RSME of DyLoc is less than 8 cm for all scenarios and the performance of DyLoc is very close in these scenarios. This can be justified by the fact that the scattering environment is very rich and the pervasiveness of paths helps DyLoc to obtain a robust location estimation to environment changes. This fact is also reflected in the RSME of PredRNN. Again, the performance of PredRNN is pretty close in the 3 scenarios and the error changes between 15 cm to 18 cm. Interestingly, it seems that the PredRNN performs better in predicting the LOS path position in the ADP rather than the NLOS paths. We may explain this phenomena by the fact that the LOS path is stronger than the NLOS paths and the predictive network could learn to predict its location better than weaker NLOS paths. Thus, centimeter accuracy in the indoor environments is achievable using DyLoc.% It should be mentioned that some recent works report centimeter accuracy in indoor environments using ultra wide-band (UWB) impulse localization methods \cite{zafari2019survey}. UWB localization techniques requires at least three fully synchronized anchor nodes while DyLoc can achieve the same accuracy-level using only a single anchor node and a narrow-band signal.}
%\KV{it is a little strange that you mention UWB here. I would have expected it in section I.} \FR{I added this qulitative comparision to bring some comparision between an available model based technique and DyLoc as one of the reviewer comments requested.

\vspace{-1mm}
\section{Conclusion}
\label{conc}
\vspace{-1mm}
We have introduced a novel framework to address data-driven localization in dynamic environments. We have discussed that proposed deep learning algorithms in the literature fail to tackle localization in dynamic environments since they are principally dependent on prolonged tasks of data gathering and network training. Taking advantage of a meaningful representation of communication channel, we have devised an algorithm to discover dynamic changes in the propagation environment. Based on that, we have developed DyLoc to perform localization in time-varing environments. We have showcased the performance of DyLoc in indoor and outdoor environments. Our results have shown that DyLoc is able to pursue localization accurately in the both environments. Moreover, simulation results have revealed that when the number of multipath increases, DyLoc becomes more robust to time-varying changes.
\vspace{-1mm}
\section*{Acknowledgment}
\vspace{-1.5mm}
This work is supported by the National Science Foundation under Grant No. CCF-1718195.
\vspace{-1.5mm}
%The current work has exhibited some preliminary results on potentials of using deep video processing techniques for tackling complex wireless communication problems. Nonetheless, the proposed architecture should be the subject of further explorations. Furthermore, the proposed perspective can be generalized to tackle other massive MIMO wireless communication problems at hand, specifically those which are bound to be tackled in dynamic environments.  Moreover, real world datasets that record both static and dynamic behaviour of communication environments would be a huge leap forward for developing robust techniques addressing highly dynamic problems.

\bibliographystyle{IEEEbib}
\bibliography{refs}
\end{document}